\documentclass[conference]{IEEEtran}
\IEEEoverridecommandlockouts

\usepackage{url}
\usepackage{tikz}
\usepackage{braket}
\usepackage{subfig}
\usepackage{xcolor}
\usepackage{amsmath}
\usepackage{balance}
\usepackage{environ}
\usepackage{physics}
\usepackage{colortbl}
\usepackage{enumitem}
\usepackage{amsfonts}
\usepackage{booktabs}
\usepackage{graphicx}
\usepackage{textcomp}
\usepackage{algorithm}
\usepackage{algpseudocode}
\usepackage[normalem]{ulem}

\newcommand{\sol}{Quorum}

\begin{document}

\title{\textbf{\sol{}}: Zero-Training Unsupervised Anomaly Detection using Quantum Autoencoders\vspace{0cm}}

\author{\IEEEauthorblockN{Jason Zev Ludmir}
\IEEEauthorblockA{
\textit{Rice University}\\
Houston, USA}
\and
\IEEEauthorblockN{Sophia Rebello}
\IEEEauthorblockA{
\textit{Rice University}\\
Houston, USA}
\and
\IEEEauthorblockN{Jacob Ruiz}
\IEEEauthorblockA{
\textit{Stanford University}\\
Stanford, USA}
\and
\IEEEauthorblockN{Tirthak Patel}
\IEEEauthorblockA{
\textit{Rice University}\\
Houston, USA
}}

\maketitle

\begin{abstract}

Detecting mission-critical anomalous events and data is a crucial challenge across various industries, including finance, healthcare, and energy. Quantum computing has recently emerged as a powerful tool for tackling several machine learning tasks, but training quantum machine learning models remains challenging, particularly due to the difficulty of gradient calculation. The challenge is even greater for anomaly detection, where unsupervised learning methods are essential to ensure practical applicability. To address these issues, we propose \sol{}, the first quantum anomaly detection framework designed for unsupervised learning that operates without requiring any training.

\end{abstract}

\pagestyle{plain}

\section{Introduction}
\label{sec:introduction}

Anomaly detection plays an essential role in various industries, from identifying fraudulent transactions in finance to detecting irregularities in power grids \cite{POURHABIBI2020113303,https://doi.org/10.1155/2022/1870136,pistoia2021quantum, 9587107}
As datasets grow in complexity, traditional machine learning (ML) methods struggle with scalability and accuracy. Quantum computing offers a promising new approach, with its potential to accelerate computations and detect subtle patterns and correlations in data. Leveraging quantum algorithms for ML tasks, particularly for anomaly detection, could transform how we tackle these challenges~\cite{preskill2023quantum,preskill2018quantum,thakkar2024improved}.

\vspace{2mm}

\noindent\textbf{The Challenge.} Applying quantum computing to anomaly detection presents significant hurdles. Quantum machine learning (QML) models typically rely on parameterized circuits that require training, which is challenging due to the complexity of computing gradients in quantum systems~\cite{liu2018quantum,tang2022recent}. Quantum systems require gradient calculations from first principles using the parameter shift rule, and these gradients are prone to exponential vanishing in ``barren plateau'' regions~\cite{herr2021anomaly}. Moreover, anomaly detection, by nature, is unsupervised, adding another layer of difficulty since no labeled data is available to guide the training~\cite{sakhnenko2022hybrid,lee2023computational}. These two factors -- quantum training complexity and unsupervised learning requirements -- create a considerable challenge for developing efficient quantum anomaly detection methods~\cite{sakhnenko2022hybrid,kukliansky2024network}. 

\vspace{2mm}

\noindent\textbf{The Gap.} \textit{Current quantum-based approaches for anomaly detection fall short because they still require training and often rely on supervised or semi-supervised learning.} These methods involve optimizing quantum circuits with labeled data, which limits their applicability in real-world scenarios where such data is scarce. This dependency on training creates both computational overheads and reduces the flexibility needed for fully unsupervised tasks~\cite{liu2018quantum,herr2021anomaly,ngairangbam2022anomaly,hdaib2023quantum,sakhnenko2022hybrid,kukliansky2024network,ahmed2016survey}.

\vspace{2mm}


\noindent\textbf{Our Solution.} To address this, we propose \sol{}, the first quantum anomaly detection framework that requires zero training and is designed for unsupervised learning. \sol{} leverages quantum principles such as amplitude encoding, random quantum transformations, and SWAP tests to identify anomalies without needing any parameter optimization~\cite{liu2018quantum,herr2021anomaly}. By utilizing quantum correlations and random projections, \sol{} detects anomalies based on deviations from the statistical norms of quantum transformations~\cite{mancilla2022preprocessing,ngairangbam2022anomaly,herr2021anomaly}.

First, \sol{}'s open-source technique carefully distributes the data into buckets based on the likelihood of anomalous events in the dataset before embedding it into quantum states using amplitude encoding~\cite{hdaib2023quantum,alcazar2020classical}. Then, \sol{} applies random quantum transformations to this data and uses a SWAP test to measure similarly between quantum states~\cite{herr2021anomaly,liu2018quantum}. \sol{} constructs an ``embarrassingly parallelizable'' ensemble of such random transformations and leverages statistical measures to identify anomalies. This approach is scalable and flexible, allowing for efficient anomaly detection without the computing cost of gradient calculation and training~\cite{sakhnenko2022hybrid,ngairangbam2022anomaly}.

\vspace{2mm}

\noindent\textbf{\sol{}'s Evaluation.} We evaluate \sol{} through extensive experiments on various datasets, including medical, industrial, and lexical data. We compare its performance to a state-of-the-art method that uses a quantum neural network (QNN) which relies on training and supervised labels. Our results show that \sol{} has a 23\% higher average F1 score over the QNN across evaluated datasets. We also provide an ablation study of how \sol{} performs with different-sized subsamples (buckets). \textit{\sol{} consistently identifies subtle anomalies that the state-of-the-art method may overlook, proving to be an effective zero-training quantum solution for unsupervised anomaly detection~\cite{sakhnenko2022hybrid,ngairangbam2022anomaly}.}


\section{Background}
\label{sec:background}

Before we present the design of \sol{}, we first provide some brief but necessary background.

\begin{figure*}[t]
    \centering
    \includegraphics[width=0.99\linewidth]{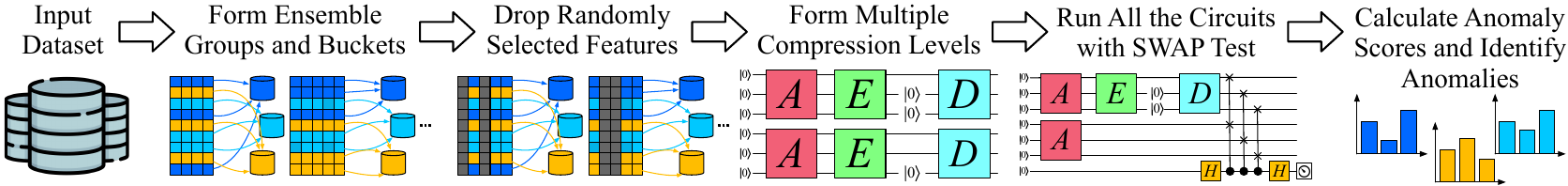}
    \caption{Visual representation of the steps taken by \sol{} to detect the anomalies in a given dataset using quantum computing.}
    \label{fig:overview}
\end{figure*}

\subsection{Quantum Computing}

The fundamental unit of quantum computation is the qubit, which exists in a superposition of binary states, represented as $\ket{\psi} = \alpha \ket{0} + \beta \ket{1}$, where $\alpha$ and $\beta$ are complex numbers that correspond to the amplitudes of the basis states $\ket{0}$ and $\ket{1}$. The probabilities of measuring the qubit in the $\ket{0}$ or $\ket{1}$ state are $|\alpha|^2$ and $|\beta|^2$, respectively, and these probabilities must sum to 1: $\norm{\alpha}^2 + \norm{\beta}^2 = 1$. Qubits can also be entangled, meaning their quantum states are correlated in such a way that they cannot be described independently of one another, which is a key property leveraged in quantum computing applications and algorithms to achieve a quantum advantage~\cite{arafath2023quantum}.

Quantum gates, represented as unitary matrices, are the basic operations applied to qubits. Common single-qubit gates include the parameterized rotation gates, such as the RX, RY, and RZ gates, which rotate a qubit around the x, y, and z axes, respectively. Two-qubit gates, such as the controlled-X (CX) gate, are used to create entanglement between qubits. These single- and two-qubit gates are defined as:
{\small
\[
RX(\theta) = \begin{pmatrix}
    \cos{\frac{\theta}{2}} & -i\sin{\frac{\theta}{2}} \\
    -i\sin{\frac{\theta}{2}} &  \cos{\frac{\theta}{2}}
\end{pmatrix}, \quad
RY(\theta) = \begin{pmatrix}
    \cos{\frac{\theta}{2}} & -\sin{\frac{\theta}{2}} \\
    \sin{\frac{\theta}{2}} &  \cos{\frac{\theta}{2}}
\end{pmatrix},
\]
\[
RZ(\theta) = \begin{pmatrix}
    e^{-i\theta/2} & 0 \\
    0 &  e^{i\theta/2}
\end{pmatrix}, \quad
CX = \begin{pmatrix}
    1 & 0 & 0 & 0 \\
    0 & 1 & 0 & 0 \\
    0 & 0 & 0 & 1 \\
    0 & 0 & 1 & 0
\end{pmatrix}
\]}
Quantum circuits are constructed by applying sequences of these gates to qubits, followed by measurements. Due to the probabilistic nature of quantum measurements, a circuit's execution yields a probability distribution of possible outcomes, which often requires many repetitions (or shots) to obtain statistically significant results \cite{https://doi.org/10.1002/cpe.6344}, \cite{10.1007/978-3-031-48421-6_25}

\subsection{Quantum Encoding and SWAP Test}

In quantum computing, classical data must be encoded into quantum states. One common method is \textit{Amplitude Encoding}, \cite{10.1049/qtc2.12032}
which maps classical data points $x_0, \dots, x_{n-1}$ to the quantum state: $\ket{\psi} = \textstyle\sum_{j=0}^{n-1} x_j \ket{j}$. This method allows $n$ classical data points to be encoded into $\log_2 n$ qubits, enabling efficient representation of high-dimensional data.

The \textit{SWAP Test} is a quantum algorithm used to determine how similar two quantum states are \cite{PhysRevLett.87.167902}, 
 \cite{PhysRevD.105.095004}. It computes the inner product $\bra{\phi}\ket{\psi}$ between two states $\ket{\phi}$ and $\ket{\psi}$. If the states are identical, the test returns a high overlap; otherwise, a lower overlap indicates dissimilarity. This is a key tool in quantum anomaly detection, where dissimilarity between two states may indicate an anomalous data point \cite{PhysRevD.105.095004}.

\subsection{Anomaly Detection}

Anomaly detection involves identifying data points that deviate significantly from the norm~\cite{wang2023quantum,gallego2022inqmad}. In unsupervised anomaly detection, we assume no labeled data is available, and the goal is to detect anomalies based solely on patterns within the data~\cite{bhowmik2024quantum,schuhmacher2023unravelling}. Common classical techniques include clustering and Isolation Forests. Clustering groups data points based on similarity, with anomalies detected as points far from cluster centroids~\cite{hdaib2023quantum,ahmed2016survey}. Isolation Forests, a tree-based algorithm, isolates anomalies by recursively splitting the data based on random feature values, where fewer splits indicate an anomaly~\cite{mensi2021alternative,marcelli2024active}.

In high-dimensional spaces, autoencoders -- neural networks trained to reconstruct their input -- are often used for anomaly detection~\cite{oyedotun2022closer,patra2022anomaly}. Autoencoders learn compressed representations of normal data, and anomalies are detected when reconstruction errors for new data points are high~\cite{ahmed2016survey,najari2022robust}. This concept extends to the quantum realm, where a quantum autoencoder can be used to compress and decompress quantum states, flagging quantum outliers based on reconstruction errors~\cite{liu2018quantum,herr2021anomaly}.

\section{Related Work}
\label{sec:related_work}

Several quantum-based approaches for anomaly detection have emerged recently, leveraging various quantum algorithms and machine learning techniques. Liu at al.~\cite{liu2018quantum} provided early evidence of the speedup and resource efficiency experienced with quantum anomaly detection. Building on this effort, Herr et al.\cite{herr2021anomaly} introduce quantum autoencoders trained with generative-adversarial networks. Both methods require structured queries or training processes, which prohibit their adaptability in unsupervised settings. Taking a hybrid quantum-classical approach, Sakhnenko et al.~\cite{sakhnenko2022hybrid} propose supervised training-based solutions that require significant quantum-classical communication. Most recently, Hdaib et al.\cite{hdaib2023quantum} propose quantum-enhanced anomaly detection with classical post-processing, still requiring circuit training.

Other works provide application-specific anomaly detection. For example, Ngairangbam et al.~\cite{ngairangbam2022anomaly} utilize quantum classifiers that rely on supervised training for high-energy physics applications. On the other, Kukliansky et al.\cite{kukliansky2024network} develop quantum anomaly detectors for network anomalies. These approaches rely on domain-specific anomaly properties and are thus not generally applicable. \textit{Thus, there is a strong need for a fully unsupervised, training-free generalized quantum anomaly detection method that leverages the unique strengths of quantum systems without the overhead of classical optimization.}

\section{Design}
\label{sec:design}

The design of \sol{} leverages quantum dynamics to identify correlations among data features and detect anomalies by comparing quantum-transformed data samples. As shown in Fig.~\ref{fig:overview}, \sol{} is structured around several key stages: preprocessing, quantum embedding, bucketing, feature selection, and circuit-based statistical analysis. Each of these steps contributes to an efficient and scalable framework for anomaly detection using quantum methods.

\subsection{Preprocessing and Normalization}
\label{subsec:preprocessing}

\sol{} begins with preprocessing the dataset, which involves a variety of steps depending on the dataset’s initial format. This typically includes transforming all non-numeric features into float values (e.g., via hashing), removing any label data that could indicate whether a sample is anomalous, and performing a range-based normalization. The normalization process is essential to ensure that all features contribute equally to the quantum state, which is critical for the subsequent quantum encoding.

Given a dataset with $M$ features, \sol{} normalizes each feature so that its maximum possible value is $\frac{1}{M}$. This ensures that the sum of the squares of all features for any sample does not exceed 1. The normalization is performed as follows:
{\small
\[
\text{normalized feature value} = \frac{\text{raw feature value for sample}}{\text{max feature value}} \times \frac{1}{M}
\]}

This normalization serves two key purposes. First, it equalizes the contribution of all features to the final quantum state, preventing any feature from dominating due to its scale. Second, it simplifies the process of amplitude embedding for quantum states, as the normalized values now range between $[0, \frac{1}{M}]$ for all of the features.

\begin{figure}[t]
    \centering
    \includegraphics[width=0.99\columnwidth]{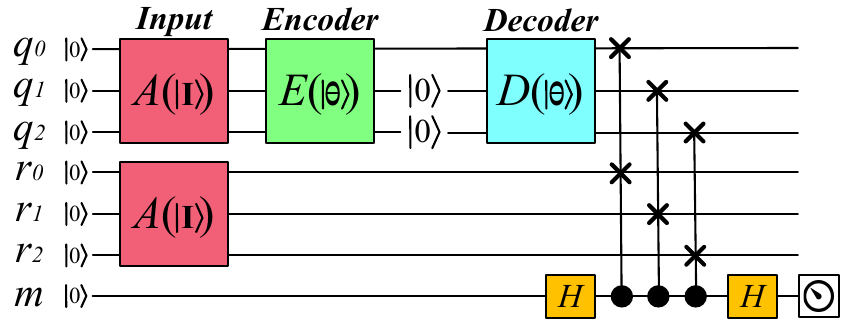}
    \caption{\sol{}'s use of SWAP test to determine the similarity between the compressed and original data sample.}
    \label{fig:swap_test}
\end{figure}

\subsection{Quantum Embedding}
\label{subsec:quantum_embedding}

After normalization, \sol{} embeds the data into quantum states using amplitude encoding. The normalized feature values are squared to convert them into probabilities, and an ``overflow state'' probability is added to account for any remaining probability mass, ensuring the total probability sums to 1. This ensures that the total probability mass of the quantum state is preserved.

A quantum circuit is then created to prepare a state vector corresponding to these probability amplitudes using amplitude embedding. This process is repeated for each data point (with $2^n$ features) using two sets of $n$ qubits within the same circuit, creating two identical encodings: one for the transformation and one as a reference (Fig.~\ref{fig:swap_test}). This dual encoding allows us to compare the transformed data with the original using a SWAP test (we'll discuss in Sec.~\ref{subsec:quantum_circuit_design} why this is required).

Each quantum circuit consists of $2n+1$ qubits, where the additional qubit serves as an ancilla for reading the SWAP test result. The SWAP test measures the similarity between the original and transformed quantum states, preserving the relative magnitudes of the features in the quantum state.

\subsection{Bucketing and Feature Selection}
\label{subsec:bucketing_feature_selection}

\begin{figure}[t]
    \centering
    \includegraphics[width=0.99\columnwidth]{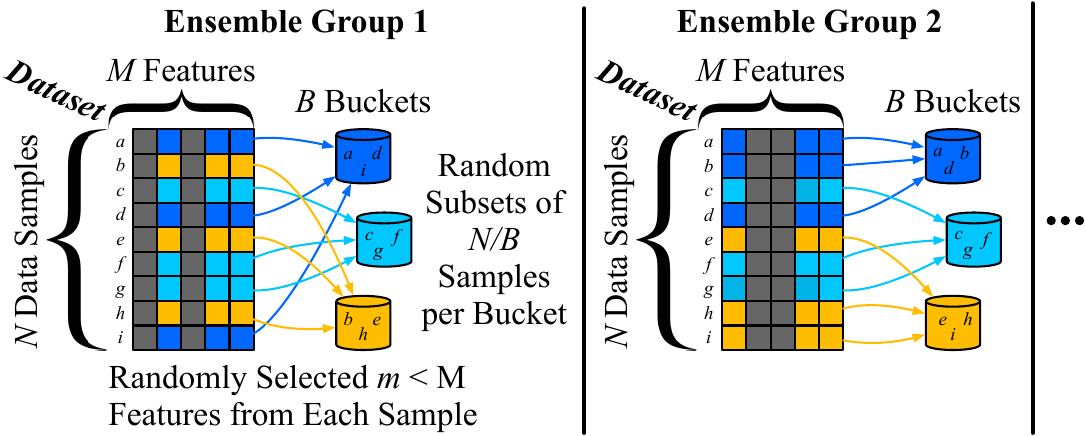}
    \caption{\sol{}'s bucketing procedure for distributing the subsampled data samples across different ensemble groups.}
    \label{fig:buckets}
\end{figure}

Following the embedding process, \sol{} employs a bucketing strategy and performs feature selection to prepare the data for anomaly detection. As shown in Fig.~\ref{fig:buckets}, the dataset is divided into a series of $B$ random subsets, or buckets. Given a dataset of size $N$, there are $N/B$ buckets. This bucketing strategy enhances the visibility of anomalies by allowing data points to be compared against smaller, more localized subsets of the dataset. The size of the buckets ($B$) is determined based on the total number of data points and the estimated proportion of anomalies, both of which determine the probability of having at least one anomaly in each bucket. By distributing the data into smaller buckets, \sol{} increases the contrast between normal and anomalous points, making it easier to detect outliers (anomalous data points).

\begin{figure}[t]
    \centering
    \includegraphics[width=0.99\columnwidth]{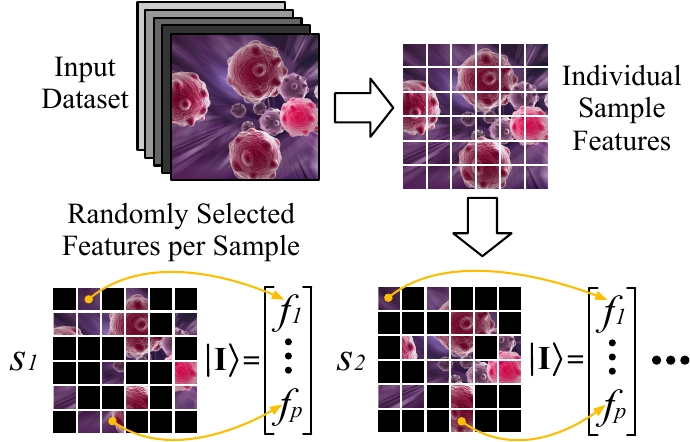}
    \caption{\sol{} subsamples features from the input space.}
    \label{fig:input_sampling}
\end{figure}

Feature selection is performed after bucketing to ensure that the data samples can fit on quantum circuits, with \sol{} using a uniform random selection strategy (Fig.~\ref{fig:input_sampling}). Unlike traditional downsampling or dimensionality reduction techniques like Principal Component Analysis (PCA), random selection offers several advantages. \textit{It is computationally faster, avoids bias towards features that might not indicate anomalies, and allows exploration of feature combinations that might otherwise be overlooked.} For each quantum circuit with $n$ qubits, \sol{} selects $m = (2^n - 1)$ features from the dataset, leaving room for an overflow state. This ensures the selected features fit within the quantum state space of $n$ qubits.

\subsection{Quantum Circuit Design and Ansatz}
\label{subsec:quantum_circuit_design}

The core of \sol{}'s anomaly detection framework lies in the design of its quantum circuits. Unlike traditional quantum autoencoders, which train parameterized gate angles to optimize encoding-decoding processes, \sol{} does not rely on learning optimal parameters. Instead, it utilizes random quantum transformations and applies statistical analysis to detect anomalies without training.

\begin{figure}[t]
    \centering
    \includegraphics[width=0.99\columnwidth]{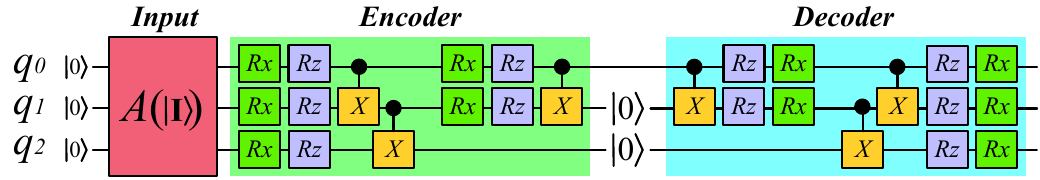}
    \caption{The ansatz leveraged by \sol{} for its autoencoder to establish correlations across input features.}
    \label{fig:ansatz}
\end{figure}

Each quantum circuit begins with an amplitude encoding of the data, as described above. The first set of encoded qubits is then passed through an ansatz, which consists of layers of RX and RZ rotations and CNOT gates. The ansatz performs random transformations on the encoded data, ensuring a high degree of variability in the quantum states (Fig.~\ref{fig:ansatz}). 

The ansatz includes three main components: an encoder circuit with randomly initialized parameters, a partial reset operation that simulates an information bottleneck by resetting a subset of qubits, and a decoder circuit that applies the inverse of the encoder. The random angles for the quantum gates in the encoder are initialized from a uniform distribution $U(0, 2\pi)$, and the decoder negates these angles to revert the transformations. A SWAP test is then performed between the transformed and original quantum states to measure their similarity. This similarity score forms the basis for detecting anomalies, as normal and anomalous data points will behave differently under random quantum transformations.

\textit{This design enables \sol{} to compress data through the autoencoder and then decode it on the other end of the reset, with the anticipation that anomalous data would be more likely to deviate from the original state when the two states are compared using the SWAP test.}

\subsection{Ensemble Groups and Statistical Analysis}
\label{subsec:ensemble_analysis}

To ensure robustness, \sol{} processes each data point through multiple ensemble groups. Each ensemble group involves randomly initialized quantum circuits ($\theta$s), ensuring that the data points are processed differently each time. Additionally, the dataset is divided into new buckets for each ensemble group, providing different perspectives on the data.

\begin{figure}[t]
    \centering
    \includegraphics[width=0.99\columnwidth]{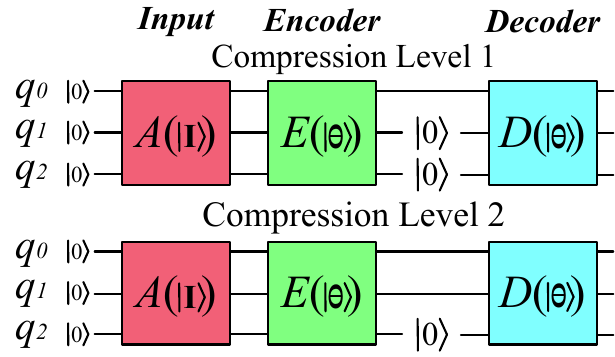}
    \caption{\sol{} leverages multiple compression levels across different ensemble groups to improve anomaly detection.}
    \label{fig:comp_level}
\end{figure}

Each ensemble group also explores multiple compression levels (Fig.~\ref{fig:comp_level}). The compression level, determined by the number of qubits reset in the partial reset operation, varies systematically from the highest compression (fewest qubits retained) to the lowest (most qubits retained). \sol{} creates a multi-dimensional view of each data point's behavior under various quantum transformations. This approach leverages the principle of random projections, where projecting data into multiple random subspaces can reveal structural information that might not be apparent in any single projection. Each ensemble group, with its unique combination of buckets, feature subsets, circuit parameters, and compression levels, represents a different ``quantum projection'' of the data.

\begin{figure}[t]
    \centering
    \includegraphics[width=0.99\columnwidth]{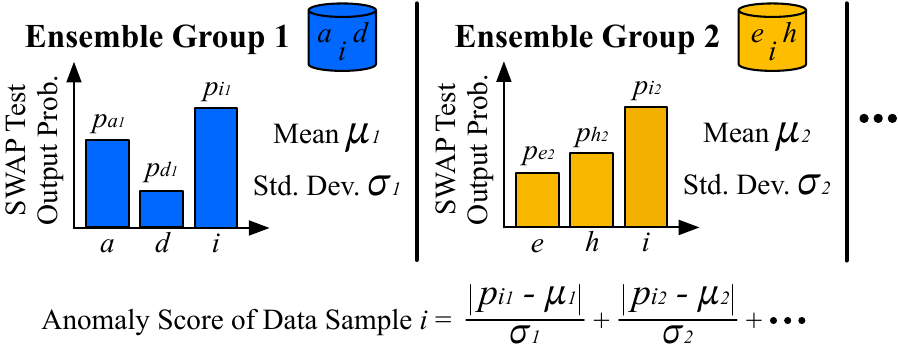}
    \caption{\sol{}'s anomaly score calculates the statistical deviation of a data sample across ensemble groups.}
    \label{fig:score}
\end{figure}

Statistical analysis is then performed on the SWAP test outputs. \textit{A key innovation in \sol{}'s approach lies in how these quantum circuit outputs are analyzed to detect anomalies.} For each bucket and each run, the mean and standard deviation of the SWAP test results are calculated. The anomaly score for each data point is derived by calculating the normalized deviation from the bucket mean divided by the standard deviation. These deviations are summed across all runs and buckets, producing an overall anomaly score for each data point (Fig.~\ref{fig:score}). A higher score indicates a greater likelihood of an anomaly.

\textit{The use of random angles in the quantum circuit is a critical aspect of \sol{}'s design.} Rather than learning to reconstruct inputs accurately, this approach creates a random projection of the data in high-dimensional Hilbert space. The statistical analysis then captures how differently each data point behaves under these random quantum transformations compared to the average behavior of points in its various buckets. 

This design allows \sol{} to detect anomalies without explicitly learning the structure of normal data or performing any optimization, instead relying on the statistical properties of how data points respond to random quantum transformations. \textit{The effectiveness of this approach lies in the fact that anomalous data points tend to behave differently under these transformations compared to normal data points, as the randomizations add more deviation to anomalous data compared to normal data.} By aggregating these behaviors across multiple random initializations, buckets, and compression levels, \sol{} aims to build a comprehensive profile of each data point's response to quantum transformations, potentially unveiling subtle anomalies that more conventional techniques might miss.

\subsection{Scalability and Flexibility}
\label{subsec:scalability}

A significant advantage of \sol{}'s design is its inherent scalability and flexibility. While our initial experiments utilized 3-qubit encodings (resulting in 7-qubit circuits), the approach can be readily scaled to accommodate larger encodings. For instance, moving to 4-qubit or higher encodings would introduce additional ``moments'' to our results for each compression level, potentially capturing even more nuanced relationships in the data. This scalability allows users to tailor the depth and complexity of the quantum transformations to their specific needs and computational resources.

Furthermore, the design of \sol{} lends itself to extensive parallelization. Each ensemble group is entirely independent of the others, making the technique embarrassingly parallel. The scalability of \sol{} extends beyond just increasing the number of qubits or ensemble groups. The flexibility in choosing the number of compression levels, the size of buckets, and the number of features selected allows users to fine-tune the balance between computational cost and the granularity of anomaly detection.

In the following section, we delve into the methodology used to obtain our evaluated results.

\section{Experimental Methodology}
\label{sec:method}

\noindent\textbf{Experimental Setup.} We evaluate \sol{} using Qiskit Aer's quantum circuit simulator. We use Python 3.10.12 and Qiskit 1.2, IBM's quantum computing language~\cite{aleksandrowiczqiskit}, which is used to run simulations of quantum circuits locally. Each quantum circuit is generated via a Qiskit QuantumCircuit object, and the quantum circuits are run through noiseless simulations. We run simulations with \sol{} and competitors on a local research cluster with Ubuntu 22.04.2 LTS on a 32-core 2.0 GHz AMD EPYC 7551P processor with 32 GB RAM.

We also perform noisy simulations, where we model our hardware after IBM's Brisbane quantum computer to provide realistic error rates. The noise parameters were obtained directly from IBM's Brisbane backend specifications; median properties are as follows: coherence times ($T_1 = 230.42~\mu\text{s}$, $T_2 = 143.41~\mu\text{s}$), gate errors (single-qubit SX gate error = $2.274\times10^{-4}$, two-qubit gate error = $2.903\times10^{-3}$), and readout error ($1.38\times10^{-2}$). Note: due to the over 100,000 runs required for \sol{}'s evaluation, it was cost-prohibitive to execute on real hardware. Thus, we use faithful simulations. 

\vspace{2mm}

\noindent\textbf{Datasets.} We evaluate \sol{} on four distinct datasets representing different anomaly detection scenarios (Table \ref{tab:datasets}). Three of the four datasets are directly derived from a related and widely-cited survey of classical unsupervised anomaly detection techniques by Goldstein and Uchida~\cite{goldsteinuchida}. The fourth dataset, which contains measurements taken from a combined cycle power plant, was taken from UCI's machine learning repository~\cite{combined_cycle_power_plant_294}. For the power plant dataset, we inserted ``plausible'' anomalies into the dataset based on ranges of values that are possible for each feature. All datasets have labels stripped for all operations until the evaluation is performed to facilitate unsupervised anomaly detection.

\begin{table}[t]
    \centering
    \caption{Datasets used for \sol{}'s evaluation. The rightmost column refers to the likelihood of at least one anomaly being placed in each bucket.}
    \scalebox{1.0}{
    \begin{tabular}{ccclp{1.7cm}}
         \textbf{Dataset} & \textbf{Samples} & \textbf{Anomalies} & \textbf{Features} & \textbf{Pr [Anomaly $\in$ Bucket]} \\
         \hline
         \hline
         Breast Cancer & 367 & 10 & 30 & 0.75 \\
         Pen-Global & 809 & 90 & 16 & 0.6 \\
         Letter & 533 & 33 & 32 & 0.95 \\
         Power Plant & 1,000 & 30 & 5 & 0.75 \\
         \hline
    \end{tabular}}
    \label{tab:datasets}
\end{table}

\vspace{2mm}

\noindent\textbf{Experimental Framework.} Each experiment consists of multiple ensemble groups, where an ensemble group represents a complete run of \sol{} with different random initializations. We use 3-qubit encodings for our primary experiments, resulting in 7-qubit circuits (including the ancilla qubit). We chose this circuit size mainly due to limitations on our computational resources as we ran simulations on our local systems. For each dataset, we execute with 1,000 ensemble members, with each member using different random bucket assignments and feature selections. We executed 4,096 shots per circuit for measurements. Increasing both shot count and ensemble members has significant impacts on performance, with benefits diminishing as they increase past a certain point. For noisy simulations, we similarly executed 4,096 shots per circuit.  We use different target probabilities for bucket size determination (see Table \ref{tab:datasets}) and provide an ablation study on the effect of bucket sizes for each of the different datasets.

\begin{figure}[t]
    \centering
    \includegraphics[width=0.99\columnwidth]{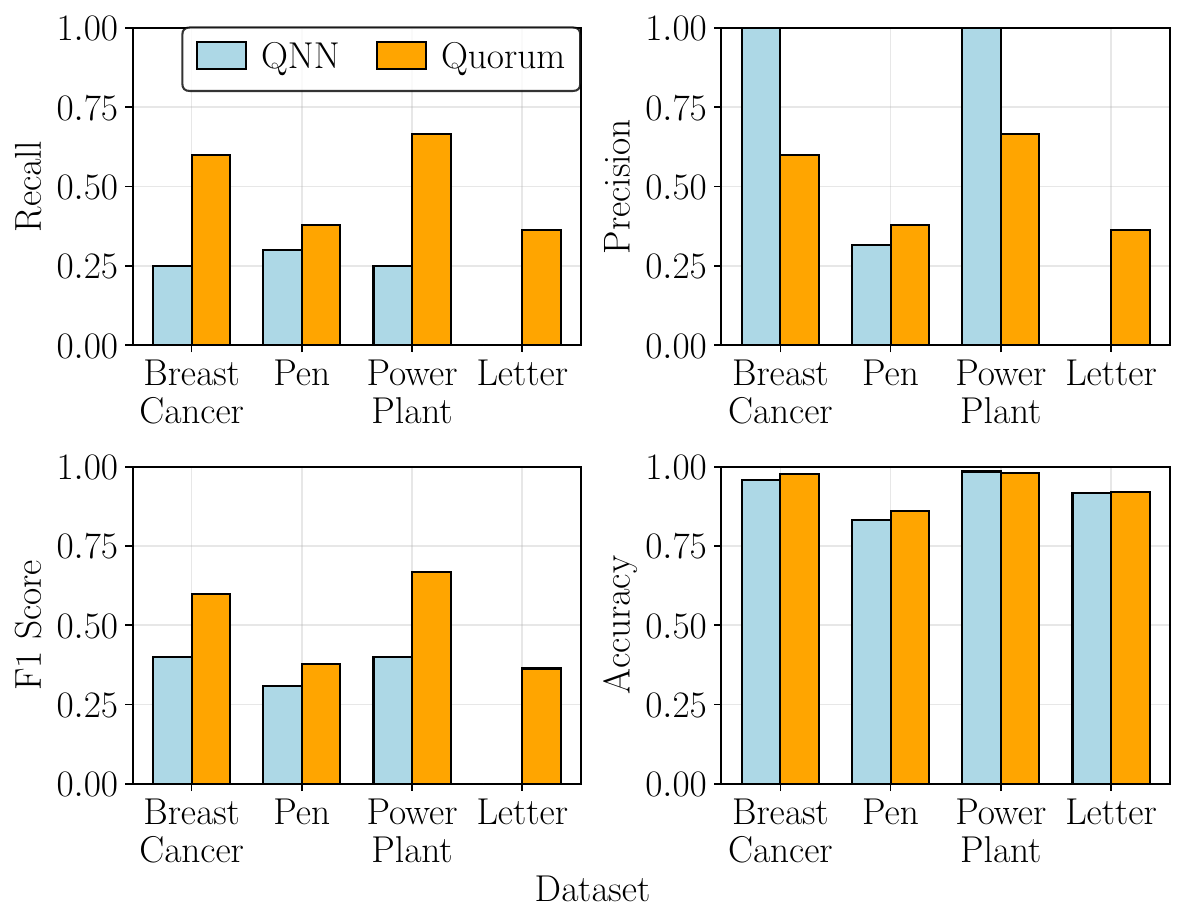}
    \caption{Comparison of performance metrics between QNN and \sol{} across four datasets. Note that the QNN did not detect any anomalies for the letter dataset, and thus, precision, recall, and F1 scores are all zero.}
    \label{fig:flagship}
\end{figure}

\vspace{2mm}

\noindent\textbf{Evaluation Metrics.} We evaluate \sol{}'s performance using several metrics: (1) \textbf{Detection Rate/Accuracy} at various percentile thresholds, measuring the fraction of true anomalies captured in the top k\% of anomaly scores. We chose to include this metric to showcase how well our technique separates anomalies from the rest of the dataset. (2) \textbf{Precision}, calculated as the ratio of correctly identified anomalies to the total number of samples flagged as anomalous. (3) \textbf{Recall}, measured as the ratio of correctly identified anomalies to the total number of true anomalies in the dataset. And (4) \textbf{F1 Score}, which is the harmonic mean of precision and recall. 

\vspace{2mm}

\noindent\textbf{Competitive Techniques.} We compare \sol{} against a state-of-the-art quantum anomaly detection technique that utilizes quantum neural networks to find anomalies in labeled datasets~\cite{kukliansky2024network} (we refer to this technique as ``QNN''). The technique is the most suitable among competitors as it shows improved performance over them and is designed for noisy quantum hardware, making it a practical benchmark. We adapted QNN for generic use since it was initially developed for network anomaly detection. Further, comparing against QNN's supervised, training-based approach helps demonstrate the advantages of our zero-training unsupervised method, particularly in scenarios where labeled data is unavailable. Note: no unsupervised, zero-training work exists to compare against as \sol{} is the first of its kind in this domain.

\section{Evaluation}
\label{sec:evaluation}

Our experimental results demonstrate the efficacy of \sol{}'s anomaly detection approach across diverse datasets. 

\vspace{2mm}

\noindent\textbf{Flagship Results.} \sol{} demonstrates balanced anomaly detection performance across all evaluated datasets -- Fig.~\ref{fig:flagship}. The recall measurements particularly highlight \sol{}'s strengths, where it consistently outperforms the QNN method. While the QNN achieves perfect precision scores on both the breast cancer and power plant datasets, this comes at the significant cost of being overly conservative in anomaly detection, leading to its notably poor recall performance. \sol{} thus achieves superior F1 performance across every dataset tested -- 23\% higher on average. \sol{}'s more nuanced approach leads to better overall detection capabilities while maintaining comparable accuracy levels across all datasets. \textit{These results demonstrate that \sol{} provides an effective approach to anomaly detection, successfully balancing precision and recall without sacrificing overall classification performance.}

\begin{figure}[t]
    \centering
    \includegraphics[width=0.99\columnwidth]{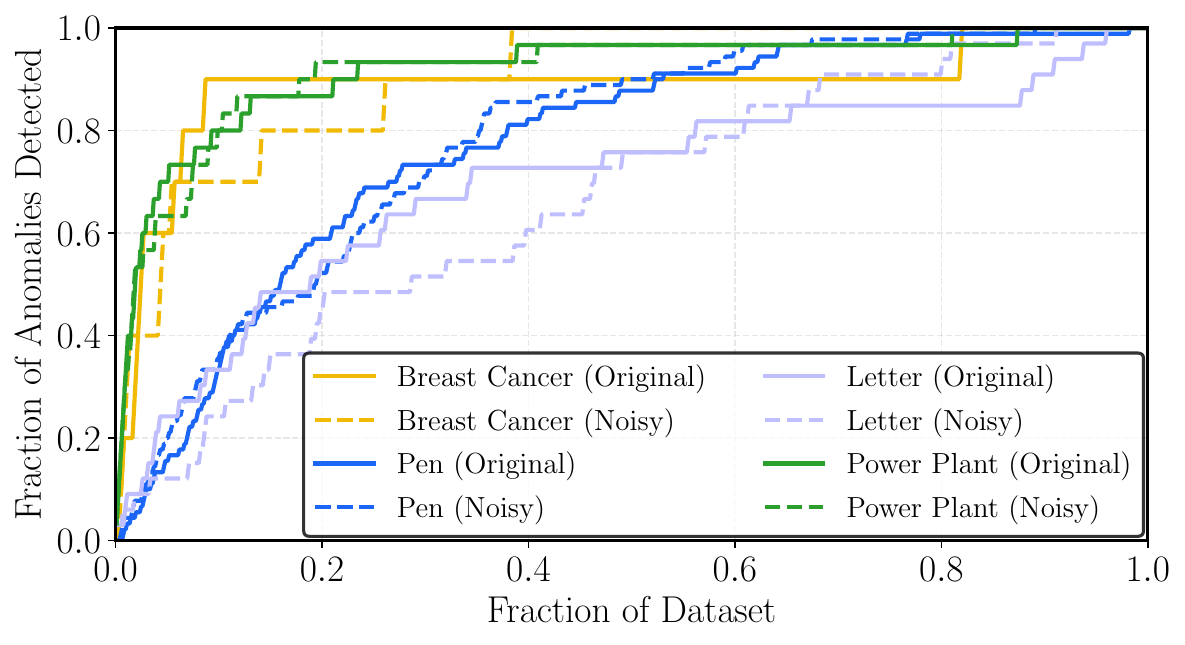}
    \caption{\sol{} groups anomalies consistently amongst data points with the highest absolute average deviations.}
    \label{fig:detection_rate_comp}
\end{figure}

\vspace{2mm}

\begin{figure}[t]
    \centering
    \includegraphics[width=0.99\columnwidth]{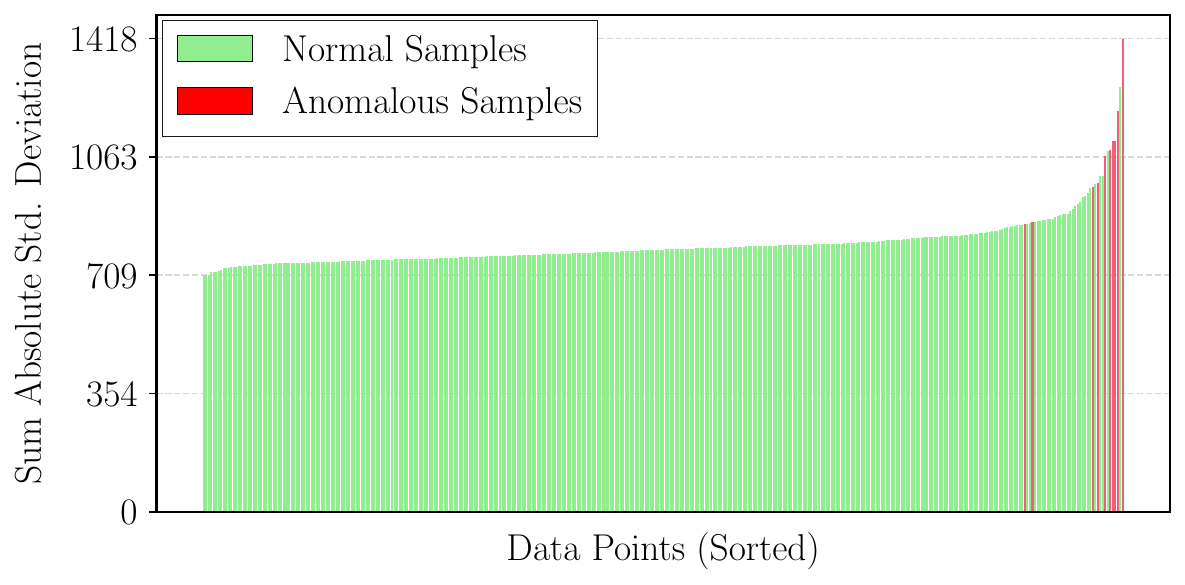}
    \caption{An example of how \sol{} separates anomalies from normal samples on the breast cancer dataset (16K shots).}
    \label{fig:bc_grouping}
\end{figure}

\noindent\textbf{Detection Rates.} In Fig. \ref{fig:detection_rate_comp}, \sol{}'s detection rate curves exhibit notably steep initial gradients, particularly for the breast cancer and power plant datasets, which achieve approximately 80\% detection rate within the first 10\% of the highest deviation samples in noiseless simulations (see Fig. \ref{fig:bc_grouping} for a detailed look at how \sol{} grouped the breast cancer data samples). The performance hierarchy among datasets likely reflects their inherent separability characteristics, with the breast cancer dataset showing the most distinctive anomaly patterns (achieving near-perfect detection at the 10th percentile), followed by power plant data.  The letter and pen datasets, while requiring a larger percentile of samples for complete detection, still maintain clear separation from random performance, achieving a roughly 60\% detection rate within the top 20\% of sample deviations for noiseless runs. When subjected to realistic quantum noise, \sol{} demonstrates a high degree of resilience. Noisy simulations closely track their noiseless counterparts across all datasets, with only minimal degradation in performance. \textit{Such inherent noise resilience is a significant advantage of \sol{} for near-term applications, as \sol{} can maintain its effectiveness even on noisy hardware without requiring high-overhead error mitigation or correction methods.}

\begin{table}[t]
\centering
\caption{F1 Scores for different bucket sizes ($p$ is the probability of at least one anomaly in a bucket).}
\scalebox{0.95}{
\begin{tabular}{l|ccccc}
\hline
Dataset & $p=0.5$ & $p=0.6$ & $p=0.75$ & $p=0.95$ & $p=0.98$ \\
\hline
Breast Cancer & 0.500 & 0.500 & 0.600 & 0.500 & 0.600 \\
Pen Digits & 0.333 & 0.389 & 0.367 & 0.389 & 0.389 \\
Letter & 0.152 & 0.182 & 0.242 & 0.273 & 0.273 \\
Power Plant & 0.600 & 0.600 & 0.633 & 0.533 & 0.600 \\
\hline
\end{tabular}}
\label{tab:bucket_f1_scores}
\end{table}

\vspace{2mm}

\noindent\textbf{Bucket Size Ablation.} Analysis of F1 scores across different bucket size configurations in Table \ref{tab:bucket_f1_scores} reveals that, as expected, very small bucket sizes generally lead to degraded performance. However, we observe that moderately sized buckets oftentimes outperform larger ones -- for instance, the letter dataset achieves its peak F1 score of 0.273 at $p=0.95$, while the breast cancer and power plant datasets show improved performance at $p=0.75$. This behavior can be explained by the trade-off between statistical significance and local sensitivity - while larger buckets provide more robust statistical estimates, smaller buckets may better capture localized anomaly patterns by reducing the ``averaging out'' effect of mixing anomalous and normal samples from the datasets.

\section{Conclusion}
\label{sec:conclusion}

In this paper, we introduced \sol{}, a novel quantum anomaly detection framework that operates without requiring any training. The framework's design, which incorporates strategic data bucketing, feature selection, and ensemble analysis, proves particularly effective at identifying anomalies while remaining computationally efficient through its inherent parallelizability. Our evaluation shows that \sol{} can achieve up to 80\% detection rate within the first 10\% of highest-deviation samples, demonstrating strong anomaly-detection power, and high resilience to noise on near term quantum systems. These results suggest that \sol{} represents a promising and wholly quantum step forward in anomaly detection.

\vspace{2mm}

\noindent\sol{}'s code is open-sourced at: {\small\texttt{\url{https://github.com/positivetechnologylab/Quorum}}}.

\section*{Acknowledgment}

This work was supported by Rice University, the Rice University George R. Brown School of Engineering and Computing, and the Rice University Department of Computer Science. This work was supported by the DOE Quantum Testbed Finder Award DE-SC0024301. This work was also supported by the Ken Kennedy Institute and Rice Quantum Initiative, which is part of the Smalley-Curl Institute.

\balance


\bibliographystyle{plain}
\bibliography{main}

\end{document}